\documentclass[letterpaper, 10 pt, conference]{ieeeconf}  

\IEEEoverridecommandlockouts                              

\usepackage{times}
\usepackage{multicol}
\usepackage{csquotes}
\usepackage[bookmarks=true]{hyperref}
\usepackage{times}
\usepackage{epsfig}
\usepackage{graphicx}
\usepackage{amsmath}
\usepackage{amssymb}
\usepackage{caption,subcaption}
\usepackage{color}
\let\labelindent\relax
\usepackage{enumitem}
\usepackage{xspace}
\usepackage[colorinlistoftodos]{todonotes}
\usepackage{algorithm}
\usepackage{algpseudocode} 
\usepackage{algorithmicx}
\usepackage{tabulary,booktabs}
\usepackage{colortbl}

\definecolor{mygrey}{gray}{0.85}

\DeclareMathOperator*{\argmin}{arg\,min}
\renewcommand{\baselinestretch}{0.995}

\makeatletter
\let\NAT@parse\undefined
\makeatother
\usepackage[numbers]{natbib}

\title{\textbf{Deep Local Trajectory Replanning and Control for 
Robot Navigation }}

\author{Ashwini Pokle$^{1}$, Roberto Mart\'in-Mart\'in$^{1}$, Patrick Goebel$^{1}$, Vincent Chow$^{1}$, Hans M. Ewald$^{1}$, \\ Junwei Yang$^{1}$, Zhenkai Wang$^{1}$, Amir Sadeghian$^{1}$, Dorsa Sadigh$^{1}$, Silvio Savarese$^{1}$, Marynel V\'azquez$^{2}$ %
\thanks{$^{1}$ Stanford University. $^{2}$ Yale University.}
}

\begin{document}

\maketitle

\begin{abstract}
We present a navigation system that combines ideas from hierarchical planning and machine learning. The system uses a traditional global planner to compute optimal paths towards a goal, and a deep local trajectory planner and velocity controller to compute motion commands. The latter components of the system adjust the behavior of the robot through attention mechanisms such that it moves towards the goal, avoids obstacles, and respects the space of nearby pedestrians. Both the structure of the proposed deep models and the use of attention mechanisms make the system's execution interpretable. Our simulation experiments suggest that the proposed architecture outperforms baselines that try to map global plan information and sensor data directly to velocity commands. In comparison to a hand-designed traditional navigation system, the proposed approach showed more consistent performance.
\end{abstract}



\section{Introduction}

\begin{figure}[t!]
    \centering
    \includegraphics[width=.95\linewidth]{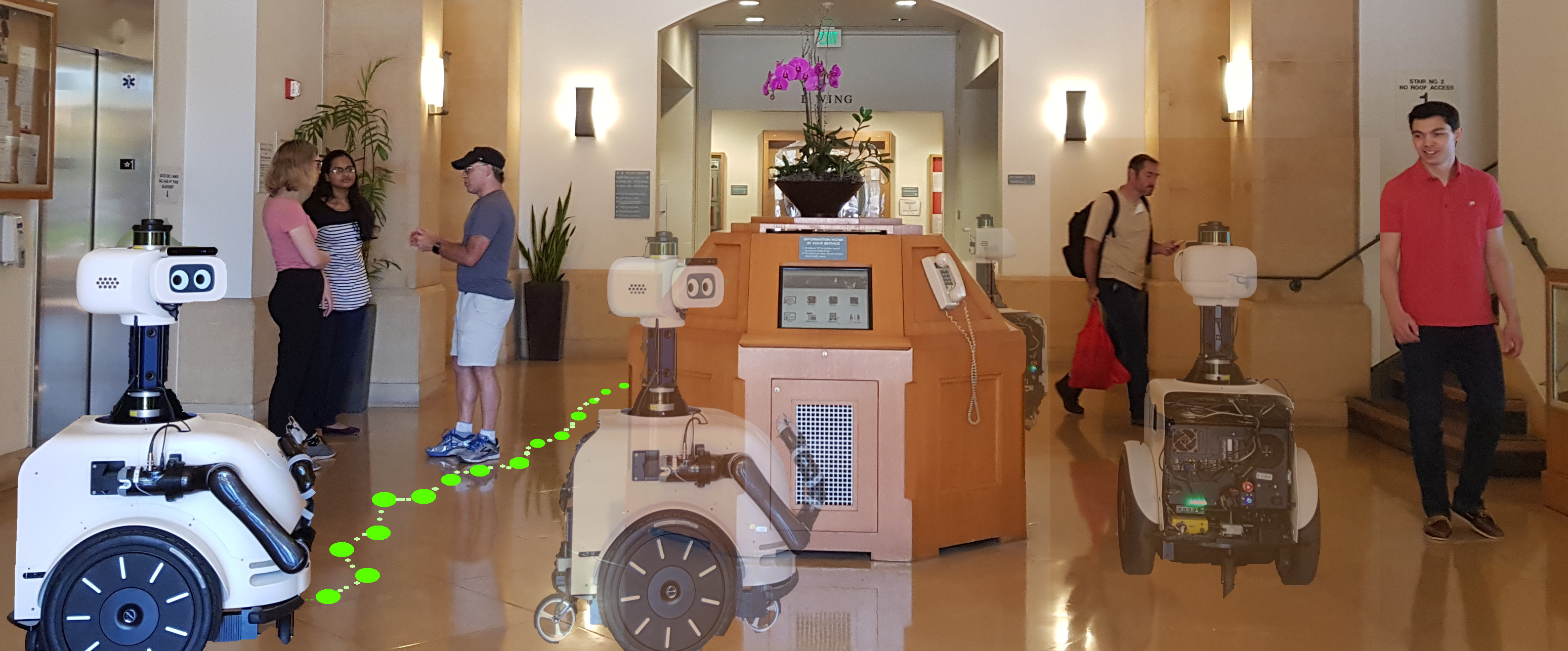}
    \caption{\small{Navigation scenario. The (solid) robot needs to reach the back of the foyer. The global planner provides an optimal solution (light green) based on a static environment map. Our navigation approach re-plans locally to adapt to the dynamic environment and control the robot towards the goal (translucent robots).}}
    \label{fig:intro}
    \vspace{-1.6em}
\end{figure}

Autonomous robot navigation in known environments encompasses 
two main problems: 1) finding a safe path for a robot to reach a desired goal location, and 2) following the path while adapting to environmental conditions ~\cite{nakhaeinia2011review}. While \textit{global planners} can efficiently find optimal motion paths \cite{lavalle2006planning}, translating these paths into robot commands -- which is traditionally the job of a \textit{reactive controller} -- can be challenging. Reactive controllers not only need to take into account kinodynamic constraints~\cite{fox1997dynamic}, but also consider plan  execution and adaptation to the environment, e.g., to avoid obstacles~\cite{khatib1986real} and respect social   conventions~\cite{kirby2010social} (Fig.~\ref{fig:intro}). 

At the core of classical approaches to reactive control is a hand-designed objective function that must balance navigation criteria to output motion commands  \cite{fox1997dynamic,gerkey2008planning,lu2013tuning}. While successful in simple situations, these approaches can be difficult to tune for dynamic human environments. One reason is that relevant criteria, like social norms, are hard to define mathematically. Even when models exist, complex interactions may emerge between model parameters and the resulting navigation behavior \cite{lu2013tuning}. Some navigation criteria may even be contradictory at times, e.g., reaching a goal in a crowded environment without violating personal space.

In this work, we combine ideas from machine learning \cite{bojarski2016end, loquercio2018dronet, pfeiffer2017perception, pmlr-v78-gao17a, sepulveda2018deep} and hierarchical planning \cite{kelly2013mobile} to improve reactive robot control. Our approach does not require hand-specifying all the parameters of the reactive controller; instead, most parameters are optimized based on example navigation data through imitation learning  \cite{pomerleau1989alvinn, muller2006off, bojarski2016end}. Similar to \cite{pmlr-v78-gao17a}, we assume that localization information is available during robot operation, and focus on studying mechanisms to combine high-level planning and learning for low-level motion control. 
%
In contrast to \cite{pmlr-v78-gao17a}, though, we use learning not only for controlling robot velocities, but also for predicting a local motion plan, which guides the velocities output by our approach. Our main insight is that by adding structure to the learning component of our system we can guide the learning process to find an appropriate complex mapping from the high-level plan to motion commands, without incurring in additional annotation costs. As suggested by our experimental evaluation, the added structure can improve overall navigation behavior in comparison to mapping a goal or a global plan  directly to commands \cite{pfeiffer2017perception, pmlr-v78-gao17a}. Predicting a local plan also facilitates system interpretability upon execution.

The proposed local trajectory planner and velocity controller of our approach adjust the behavior of the robot through simple, deep attention mechanisms. These mechanisms enable the robot to dynamically focus on different tasks: obstacle avoidance, social interaction, or following navigation plans. As a result, the robot is able to account for changing elements in the environment. It behaves in a manner that resembles complex rules that govern human use of space \cite{hall1966hidden,kendon1990conducting}. 

In summary, this work has several main contributions. First, we introduce a new system for autonomous navigation which combines planning and learning. 
The system's learning component predicts a local plan and motion commands. 
Second, we propose an attention mechanism for multimodal data fusion. 
Finally, we conduct controlled experiments on a simulated platform to evaluate the proposed system. 

\section{Related Work}
\label{sec:related_work}

Autonomous navigation has long been studied within robotics. Early navigation methods focused on path planning \cite{kunchev2006path}, while more recent approaches tend to leverage machine learning to make navigation systems less brittle to new environmental conditions \cite{thrun2005probabilistic, otte2015survey}. In some cases, machine learning is used to create systems that improve as they explore more of the environment, e.g., with reinforcement learning \cite{kober2013reinforcement} or via knowledge transfer in lifelong learning settings \cite{thrun1995lifelong}. In other cases, machine learning is used to model preferences for the navigation task with the help of a teacher \cite{argall2009survey}. For instance, prior approaches have used structured prediction  \cite{bagnell2007boosting} or inverse reinforcement learning \cite{ziebart2008maximum} to find cost functions that encode preferences for navigating through parts of an environment. Closer to our work, some methods have focused on directly learning motion policies with imitation learning, e.g., \cite{pomerleau1989alvinn, muller2006off, bojarski2016end}. As discussed in \cite{ho2016model}, directly mapping states to actions can be more efficient than learning a cost function for imitation. 
Due to limited space, we encourage readers interested in more details of these different approaches to refer to \cite{kunchev2006path,otte2015survey,kober2013reinforcement,argall2009survey, bagnell2015invitation}.

Similar to \cite{bojarski2016end, loquercio2018dronet, pfeiffer2017perception, sepulveda2018deep}, we use deep learning \cite{Goodfellow-et-al-2016} to parameterize a motion policy. This approach allows us to forgo hand-engineered features for sensor data. In contrast to most of these efforts, though, we do not aim to solve the whole navigation problem with a single function trained in an end-to-end fashion. Instead, we leverage planning in conjunction with deep learning for autonomous navigation. 

Inspired by IntentionNet \cite{pmlr-v78-gao17a}, we use a global planner to solve for the general direction that a robot should follow to reach a desired destination in a known environment, as well as use deep learning for motion control. But different to \cite{pmlr-v78-gao17a}, (1) our approach explicitly considers the presence of people nearby the robot, (2) processes raw lidar measurements instead of RGB images, (3) represents global plans via trajectories, and (4) enforces additional structure on the learning component of the navigation system.
%
%
%
Our rationale behind these considerations are as follows. First, providing information about people's motion directly to our navigation system facilitates interactions in human environments. Second, providing raw lidar measurements to our system reduces the complexity of the input space and facilitates system development through simulation in comparison to using raw images. Lidar can also help with obstacle avoidance, as in \cite{pfeiffer2017perception}, because it measures depth directly and typically has a wider field of view than cameras. Third, representing global plans via trajectories, instead of rendering them on maps, reduces even further the dimensionality of the input space. Fourth, adding structure to the learning component of our navigation system offers an opportunity to add supervision and facilitate interpretability. Our approach does not aim to build an environment map as the robot navigates \cite{gupta2017cognitive}, but assumes that a map is given for high-level, global planning.  

The study of how people use and share space, or proxemics \cite{hall1966hidden}, is relevant for social robot navigation, e.g., see \cite{rios2015proxemics} for a recent review. Generally, most methods for social navigation explicitly model interactions among agents or social conventions \cite{kirby2010social,henry2010learning,lu2013tuning,kollmitz2015time,luber2012socially,trautman2013robot,kretzschmar2016socially,pfeiffer2016predicting, ChenELH17,okal2016learning,johnson2018socially}. Other methods, like ours, implicitly model these aspects and let navigation strategies emerge through imitation  \cite{pmlr-v78-gao17a,shiarlis2017acquiring,tai2018social}. 


We leverage simulations to facilitate training motion policies and conduct system evaluations. Similar to work by \citet{tai2018social} and \citet{koltun-policy-transfer:18}, we avoid providing complex sensory input, like images, to our method. Instead, our deep  model takes as input lidar  and information derived from complex signals, like people's position nearby the robot. 


\section{Approach}
\label{sec:approach}

\begin{figure*}[t]
    \centering
    \includegraphics[width=\linewidth]{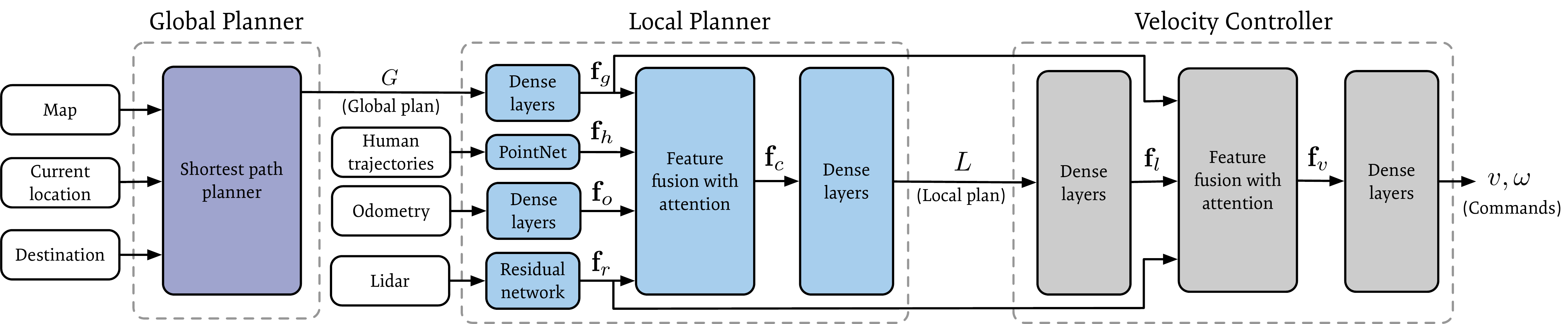}
    \caption{Proposed approach. The white boxes denote inputs to our system. The purple box, blue boxes, and gray boxes belong to the global planner, the local planner, and the velocity controller, respectively. See the text for more details. 
    }
    \label{fig:sysdiag}
    \vspace{-1em}
\end{figure*}

Building on a long tradition in robotics \cite{shiller1991dynamic,laumond1994motion,kolter2008control,peng2017deeploco,pmlr-v78-gao17a}, we use a hierarchical approach for  locomotion. Our particular hierarchy is composed of three levels: Global Planning, Local Planning, and Velocity Control. 
%
%
As shown in Fig.~\ref{fig:sysdiag}, the \textit{Global Planner} takes as input a 2D map of the environment with static obstacles, e.g., walls, the current location of the robot, and the coordinates of a desired destination (or goal) in the map. Based on these inputs, the global planner computes a path in the map for the robot to reach the desired goal location. 
%
%
The \textit{Local Planner} then focuses on a shorter time  
horizon: what are the immediate navigation steps that the robot should take to follow the global plan in an acceptable manner given observed environmental conditions? These conditions include dynamic elements, such as people, as well as static obstacles sensed in the vicinity of the robot. Finally, the \textit{Velocity Controller} commands motions to the robot based on the navigation plans.



Explicitly or manually defining socially appropriate navigation behavior is hard. Human behavior is malleable and social norms often change from one social context to another \cite{hall1966hidden, kendon1990conducting}. Thus, we resort to imitation learning to model appropriate behavior. Our main assumption in this work is that expert demonstrations are available to learn what constitutes good navigation patterns for a robot.


We implement the  Local Planner and the Velocity Controller of our navigation approach as Neural Networks and optimize their parameters based on expert motion data. While it would have been possible to also implement the Global Planner as a differentiable function \cite{tamar2016value} and optimize for the whole system in a complete end-to-end fashion, we opt for combining machine learning with traditional planning in this work. This combination aims to take advantage of the benefits of both approaches, while keeping the implementation of our system practical. We use planning because it is fast and reliable for computing obstacle-free routes in static environments, and we use deep networks because they have a great capacity for reasoning about raw sensor data and modeling complex phenomena like social behavior \cite{sadeghian2017tracking,Vemula-2018-107260}. Note that the learning component of our hierarchical navigation approach has more structure than similar prior work \cite{pfeiffer2017perception,pmlr-v78-gao17a}. We use a differentiable Local Planner to provide additional supervision to the learning component of our system and  facilitate interpreting its execution.


\subsection{Global Planning: Finding a Route to the Goal}
\label{ssec:global_planning}
As a first step when navigation starts, our system computes a collision free path from the initial location of the robot $\mathbf{x}_0$ to the desired goal location $\mathbf{g}$ in a static map of the environment. The path is computed with Dijkstra's shortest path algorithm \cite{dijkstra1959note} and saved as a reference during navigation. 

The Global Planner of our system provides a down-sampled version of the full global plan to the Local Planner. Down-sampling preserves the desired motion direction and highlights the relevant part of the global plan for successive processing. Let the output of the Global Planner be a down-sampled path $G=\{(x_i,y_i)\,|\,1 \leq i \leq 10\}$, with each $(x_i,y_i)$ a waypoint belonging to the original full plan. We enforce that the first waypoint $(x_1,y_1)$ in $G$ is the closest waypoint in the full plan to the current location of the robot, and that the remaining waypoints are at least 0.5 meters from the previous one in $G$. If not enough waypoints that satisfy these constraints remain in the full plan, then the goal is used instead.

\subsection{Local Planning: Predicting Navigation Subgoals}
\label{ssec:local_planning} 

Let the inputs to the local planner be the down-sampled global plan $G$, the lidar range measurements $R = \{r_i\,|\, 1 \leq i \leq M \}$, the robot's linear and angular velocities $O = (\hat{v}, \hat{\omega})$ from odometry, and a set of observed human trajectories $H = \{ \mathcal{T}_i \,|\, 1 \leq i \leq P \}$  with $\mathcal{T}_i = \{(x_j,y_j)\,|\,t-k \leq j \leq t\}$ the 2D coordinates of person $i$ for the last $k$ time-steps relative to the current pose of the robot. Features for these inputs are computed by:

\begin{description}[align=left,leftmargin=0em,labelsep=0.2em,font=\textbf]
\item[Global Plan Features.] The model flattens $G$ into $g \in \mathbb{R}^{20}$, and projects $g$ into a higher dimensional space through two fully connected ReLU layers with 32 and 512 units, respectively. The result is a feature vector $\mathbf{f}_g \in \mathbb{R}^{512}$.

\item[Lidar Features.] The model uses the residual network from \cite{pfeiffer2017perception} to compute the lidar features $\mathbf{f}_r \in \mathbb{R}^{512}$. The residual network is composed of 5 convolutional layers and two residual connections. See Fig. 2 in \cite{pfeiffer2017perception} for more details.

\item[Odometry Features.] The models projects $O$ into a higher dimensional space through two fully connected ReLU layers, similar to the Global Plan features. The result is a feature vector $\mathbf{f}_o \in \mathbb{R}^{512}$.

\item[Human Trajectory Features.] In the spirit of \cite{gupta2018social}, our model uses a PointNet network \cite{qi2017pointnet} to compute features for a set of past human trajectories. This choice frees us from needing to specify an order in which human coordinates should be input to our model. More specifically, the local planner organizes observed human trajectories $H$ into a matrix $H' \in \mathbb{R}^{P \times 2k}$, where each row corresponds to the data of one person. The rows are independently projected into a higher dimensional space of 1024 dimensions through a three layer perceptron with ReLU activations. 
The resulting matrix in $\mathbb{R}^{P \times 1024}$ is then applied a max pooling operation and flattened to convert it into a vector in $ \mathbb{R}^{1024}$ that encodes information about the motion of the people nearby the robot. This data is finally transformed by a fully connected layer with ReLU activation into the feature vector $\mathbf{f}_h \in \mathbb{R}^{512}$, which has the same dimensionality as the other 
features.
\end{description}

In initial experiments, we considered combining the above features through concatenation \cite{pfeiffer2016predicting, pmlr-v78-gao17a} or outer product operations \cite{fukui2016multimodal,ben2017mutan}. We found, however, that it was hard for the robot to pay attention to multiple sources of information with concatenation. 
For example, the robot would follow the global plan and ignore obstacles observed through its lidar in some cases. We also found that, as in \cite{fukui2016multimodal}, the outer product led to very high-dimensional features that made training prohibitively expensive or prone to over-fitting.

\noindent
\textbf{Fusion of Input Features.} Inspired by prior work in text translation \cite{bahdanau2014neural}, we propose to fuse input features with a simple concatenation-based attention mechanism:
\begin{equation}
\mathbf{f}_c = [a_g\mathbf{f}_g\  a_r\mathbf{f}_r\ a_h\mathbf{f}_h\ a_o\mathbf{f}_o]
\label{eq:att}
\end{equation}
with $\mathbf{a} = [a_g\ a_r\ a_h\ a_o]^T$ coefficients that reflect the relative importance of the features. We compute the attention coefficients $\mathbf{a}$ by first concatenating the input features and transforming them into a higher dimensional space, $\mathbf{u} = \max(0,W_1\text{concat}(\mathbf{f}_g, \mathbf{f}_r, \mathbf{f}_h, \mathbf{f}_o))$ with $W_1 \in \mathbb{R}^{128 \times 2048}$ a parameter matrix. Then, the desired coefficients are given by $\mathbf{a} = \text{softmax}(W_2\mathbf{u})$ with $W_2 \in \mathbb{R}^{4 \times 128}$ trainable parameters. 

The Local Planner finally generates a plan $L$ by transforming the combined features $\mathbf{f}_c$ through 3 fully connected layers with 512, 256 and 64 units and ReLU activations, and one final linear transformation. The plan is an ordered set of poses $L = \{(x_i,y_i,\cos(\theta_i), \sin(\theta_i))\,|\,1 \leq i \leq 5 \}$ sampled at 1Hz that represent a time-dependent trajectory that the robot could follow to adapt to the dynamic environment within the next 5 seconds. $L$ guides the robot in the direction of the global plan, and encodes temporal information to enable variations in speed as necessary while navigating. 


\subsection{Velocity Control: Predicting Low-Level Commands}
\label{ssec:velocity_control}

We provide the Velocity Controller three inputs: $L$, the lidar features $\mathbf{f}_r$, and the global plan features $\mathbf{f}_g$ computed during local planning. While $L$ might suffice in cases where the prediction of the local plan is appropriate for the current environmental conditions, there might be cases when $L$ is problematic, e.g., bringing the robot close to collision due to a prediction error, or getting the robot stuck due to local minima. In these cases, range information and the global plan can help the robot avoid obstacles in close proximity and navigate towards the goal.

The Velocity Controller first computes features for the local plan $L$ by projecting it to a higher dimensional space through two fully connected ReLU layers with $32$ and $512$ units, respectively. The result are local plan features $\mathbf{f}_l \in \mathbb{R}^{512}$ with the same dimensionality of features $\mathbf{f}_r$ and $\mathbf{f}_g$.

The Controller combines features with attention, as in eq. (\ref{eq:att}), except that only three attention coefficients are needed in this case. Let these coefficients be denoted by $b_l$, $b_r$, $b_g$. Then, the combined features $\mathbf{f}_v = [b_l\mathbf{f}_l\  b_r\mathbf{f}_r\  b_g\mathbf{f}_g\ ]$.

Finally, the Velocity Controller projects the combined features $\mathbf{f}_v$ to a 2D space through 2 fully connected ReLU layers with 512 and 128 units, followed by a fully connected ReLU layer and a linear activation function. The two outputs are the linear velocity ($v$) and angular velocity ($\omega$) that command the robot to move as desired, as shown in Fig.~\ref{fig:sysdiag}.

\subsection{Learning a Local Planner and Velocity Controller}

The joint network of the Local Planner and the Velocity Controller can be seen as a policy that maps states into actions based on examples from an expert. The state is composed of the inputs to the Local Planner; the actions are the low-level, velocity commands that are output by our navigation system. Under this formulation, our objective is to compute a policy $\hat{\pi}$ that minimizes the expected loss $\ell$ with respect to the expert policy $\pi^*$ under the distribution of states $s \sim d_{\pi^*}$ induced by the expert:
\begin{equation}
    \hat{\pi} = \argmin_{\pi \in \Pi} \mathbb{E}_{s \sim d_{\pi^*}}[\ell(s,\pi)]
    \label{eq:policy}
\end{equation}

This particular view of the imitation problem follows close related work \cite{pfeiffer2017perception,pmlr-v78-gao17a} and is typically known as Behavioral Cloning \cite{pomerleau1989alvinn, bojarski2016end,shiarlis2017acquiring}.\footnote{It is  possible to frame the objective as finding a policy that minimizes the loss $\ell$ under its distribution of states $s \sim d_{\pi}$  \cite{ross2011reduction}. Even though this is computationally more expensive, prior work suggests that it can improve imitation \cite{Pan-RSS-18,venkatraman2016improved}. This alternative formulation is interesting future work.
} Our main assumption is that expert motion trajectories are available for supervised learning.

We learn the parameters of the Neural Networks of our system through back-propagation with three Adam optimization procedures \cite{adamopt} aimed at providing an easier path to learning, similar to curriculum learning \cite{bengio2009curriculum} or shaping \cite{krueger2009flexible}. Unless otherwise noted, we use L2 loss with a batch size of 300 samples. We pick the best parameters after 100 epochs.

\noindent
\textbf{Training Procedure.} First, we compute lidar features $\mathbf{f}_r$ with the help of the navigation model proposed by \citet{pfeiffer2017perception}. We train their model by minimizing the loss on robot velocity. We then load the pre-trained weights for $\mathbf{f}_r$ into the Residual Network of our Local Planner (Fig.~\ref{fig:sysdiag}) and optimize for the rest of this component. Our insight is that we can use expert motion trajectories to supervise the Local Planner without collecting additional data. Our ground truth for the local plan $L$ at a given time-step $t$ is the pose of the robot at $t+1$s, $t+2$s, etc., in the future, which is readily available after collecting expert trajectories. Finally, we load the Local Planner into a network with all of the learning components of our navigation system, and train the Velocity Controller.



\section{Experimental Setup}
\label{sec:expsetup}


\begin{description}[align=left,leftmargin=0em,labelsep=0.2em,font=\textbf,itemsep=0.5em]
\item[Experimental Platform.] We use a simulated JackRabbot 2 robot for data collection and evaluation. The robot is a differential-drive mobile manipulator with a forward-facing 2D SICK lidar and an Occam $360^\circ$ stereo camera, among other sensors. JackRabbot's software stack uses the \href{http://www.ros.org}{Robot Operating System (ROS)} \cite{quigley2009ros} for inter-process communication and logging, 
OpenSlam's Gmapping for creating environment maps \cite{grisettiyz2005improving}, ROS's Adaptive Monte Carlo Localization algorithm for pose estimation \cite{fox2002kld}, and YOLO \cite{redmon2017yolo9000} and DeepSort \cite{Wojke2017simple} for visually detecting and tracking pedestrians all around. Pedestrian tracks are converted from 2D to 3D with calibrated cameras and lidars.  

\item[Simulation Environments.] We use the \href{http://gazebosim.org/}{Gazebo} simulator for our experiments. To test navigation algorithms on the simulated robot, we use the robot's true pose from Gazebo instead of its localization algorithm, the true pedestrian motion relative to the robot instead of its vision pipeline, and a simulated forward-facing lidar. These changes allowed us to systematically study robot behavior without potential confounds due to perception errors. 

To evaluate navigation in varied scenarios, we added obstacles to the simulation, e.g., 0.5-0.75\,m$^3$ boxes and cylinders, and implemented a Social Forces model  \cite{helbing1995social} for the pedestrians in Gazebo. The Social Forces model enables each pedestrian to adapt their path to reach navigation sub-goals.

Overall, we consider 7 different environmental conditions in our experiments (Table \ref{tab:env}). Six conditions (E1-E6) use 
the same map of the foyer of a university building, which is depicted in Fig. \ref{fig:intro}. The other condition uses the map of a laboratory environment. We use four different environmental conditions for training or tuning parameters (E1-E4) and four for testing (E4-E7). In particular, we use E5 and E6 for testing generalization to new environmental conditions in a known map, and E7 for testing generalization to a new environment. Each \textit{scenario} indicated under the ``\# Train'' and ``\# Test'' columns of Table \ref{tab:env} corresponds to a different pair of start-end locations for the robot, and navigation sub-goals for the pedestrians. The scenarios used for testing in E4 are different than those used for training or tuning methods.

\begin{table}[t]
\centering
\caption{\label{tab:env} \small{Configuration of simulation environments. ``G'' denotes Geometric obstacles, ``SP'' is Static People, and ``MP'' is Moving Pedestrians. ``\# Train '' indicates the number of scenarios from the corresponding env. configuration that were used for training models. ``\# Test'' indicates the number of (new) scenarios used for testing.}}
\small{
\begin{tabular}{c|cccccc}
\bf Env. & \bf Map & \bf{\# G} & \bf{\# SP} & \bf{\# MP} & \bf \# Train  & \bf{\# Test} \\ \hline
\arrayrulecolor{mygrey}
E1 & Foyer & 8 & 0 & 0 & 42 & 0\\ \hline
E2 & Foyer & 12 & 0 & 0 & 17 & 0\\ \hline
E3 & Foyer & 6 & 3 & 2 & 23 & 0\\ \hline
E4 & Foyer & 3 & 2 & 3 & 1364 & 100\\ \hline
E5 & Foyer & 1 & 1 & 1 & 0 & 100 \\ \hline
E6 & Foyer & 8 & 3 & 3 & 0 & 50\\ \hline
E7 & Lab. & 8 & 3 & 3 & 0 & 50
\end{tabular}
}
\vspace{-2em}
\end{table}

\item[Data Collection.] We collect a dataset of 1446 expert motion trajectories -- each corresponding to a different scenario -- for the robot with Gazebo. The robot was tele-operated with a gamepad controller for 600 scenarios in E1-E4. The remaining 764 expert motion trajectories were selected from a set of runs of the ROS' Navigation Stack with social costs enabled \cite{lu2014layered} in E4, as tele-operation is time consuming and deep learning benefits from big amounts of data. We tried various approaches for combining tele-operated and auto-generated data, e.g., transfer learning, but found that using all of it for training at once led to best performance in general.

\item[Baselines.] We compare our system with three baselines:
\begin{description}[align=left,leftmargin=0em,labelsep=0.2em,font=\normalfont\textit,itemsep=0.2em]
\item [- Nav. Stack.] ROS' layered Navigation Stack with social costs \cite{lu2014layered}, Dijkstra's algorithm for global planning, and an Elastic Band local controller \cite{quinlan1993elastic}.
\item [- Goal Controller (GC).] Deep controller that maps lidar range measurements and a 2D navigation sub-goal to velocity commands \cite{pfeiffer2017perception}. The sub-goal moves along the global plan. It is generally 2\,m ahead of the robot until JackRabbot approaches the destination and the sub-goal becomes the goal. Input features are fused with concatenation as in \cite{pfeiffer2017perception}.  
\item[- Trajectory Controller (TC).] Deep controller that maps lidar measurements and the down-sampled global plan $G$ to velocity commands. This model is based on \cite{pmlr-v78-gao17a} but takes lower-dimensional inputs, which facilitates learning. Input features are combined with concatenation as in \cite{pmlr-v78-gao17a}. 
\end{description}
The GC and TC baselines can be considered ablation models of the proposed deep local trajectory planner and velocity controller. These baselines use the same lidar features $\mathbf{f}_r$ and global plan features $\mathbf{f}_g$ as the proposed system.

\item[Objective Metrics.] For a given set of test scenarios, we consider the following metrics:
\begin{description}[align=left,leftmargin=0em,labelsep=0.2em,font=\normalfont\textit,itemsep=0.2em]
\item [- Running Time.] Total running time for all the scenarios.
\item [- Distance.] Total distance that the robot traversed considering all the scenarios.
\item [- Linear Vel.] Average linear velocity that the robot reported considering all the scenarios.
\item [- Reached Goal (RG).] Percentage of the scenarios in which the robot reached the goal.
\item[- Failure (F).] Percentage of the scenarios in which the robot failed catastrophically and tipped over, e.g., after a collision. 
\item[- Collisions or Near Collisions (C).] Number of events in which the robot's lidar sensed an obstacle closer than 0.3\,m. The count considers all scenarios; it is not averaged. 
\item[- Pedestrian Collisions (PC).] Number of events in which the robot collided with a pedestrian (their distance was less than 0.5 m). The count considers all scenarios.
\item[- Violations of Personal Space (PS).] Number of events in which the robot violated personal space \cite{hall1966hidden} and was less than 1.2 m from a person. The count considers all scenarios. 
\end{description}

\item[Subjective Metrics.] We conduct a survey to gather qualitative perceptions of navigation performance. Participants rate navigation behavior on three 5-point Likert scales: \textit{Aggressive}, \textit{Natural}, and \textit{Efficient} navigation.

\item[Other Details.] We consider that the robot reached the goal if its distance to the destination is less than 0.5\,m. For all the scenarios, we limit the time that the robot can take to reach the destination to 1.5\,min. Finally, the robot has a safe, maximum limit of 0.6\,m/s and 1.2\,rad/s on its absolute linear and angular speeds.

\end{description}





\section{Experimental Evaluation}
\label{sec:expev}

\begin{table*}[t]
\centering
\caption{\label{tab:results} Objective results. $\uparrow$: higher is better; $\downarrow$: lower is better. Please see Sec. \ref{sec:expsetup} and \ref{sec:expev} for more details. }
\small{
\begin{tabular}{c|c|cccccccc}
\bf Model & \bf Env. & \bf Running Time & \bf Distance & \bf Linear Vel. & \bf RG $\uparrow $ & \bf F $\downarrow $ & \bf C$\downarrow $ & \bf PC$\downarrow $ & \bf PS $\downarrow $ \\ \hline
\arrayrulecolor{mygrey}
Nav. Stack & E4 & 60.1 min & 802.7 m & 0.22 m/s & 91\% & \bf 0/100 & \bf 64 & \bf 1 & 82 \\ \hline
Goal Cont. (GC) & E4 & 55.4\,min & 901.9\,m & 0.27\,m/s & 89\% & 2/100 & 167 & 12 & 108\\ \hline
Traj. Cont. (TC)  & E4 &  49.8 min & 993 m & 0.33 m/s & 85\% & 7/100 & 129 & 7 & 112 \\ \hline
Proposed System  & E4 & 54.2\,min & 923.3\,m & 0.28\,m/s & \bf 95\% &\bf  0/100 & 102 & \bf 1 & \bf 76 \\ \hline
\arrayrulecolor{black}\hline \hline
\arrayrulecolor{mygrey}
Nav. Stack & E5 & 56.8\,min & 892.2\,m & 0.26\,m/s & \bf 98\% & \bf 0/100 & \bf 0 & 1 & 28 \\ \hline
Goal Cont. (GC) & E5 & 46.4\,min & 930.2\,m & 0.33\,m/s & \bf 98\% & \bf 0/100 & 52 & 1 & \bf 27  \\ \hline
Traj. Cont. (TC)  & E5 & 45.3\,min & 961.5\,m & 0.35\,m/s & 97\% & 1/100 & 61 & \bf 0 & 43 \\ \hline
Proposed System  & E5 & 59.6\,min & 941.7\,m & 0.26\,m/s & 93\% & \bf 0/100 & 27 &1 &34 \\ \hline
\arrayrulecolor{black}\hline \hline
\arrayrulecolor{mygrey}
Nav. Stack & E6 & 38.7\,min & 424.8\,m & 0.18\,m/s & 82\% & \bf 0/50 &  \bf 67 & \bf 0 & 80 \\ \hline
Goal Cont. (GC) & E6 & 34.0\,min & 333.0\,m & 0.16\,m/s & 74\% & 1/50 & 178 & 1 & 65 \\ \hline
Traj. Cont. (TC)  & E6 & 39.4\,min & 472.2\,m & 0.2\,m/s & 66\% & 1/50 & 152 & 4 & 65 \\ \hline
Proposed System  & E6 & 34.4\,min & 484.4\,m & 0.23\,m/s & \bf 98\% & \bf 0/50 & 112 & 1 & \bf 62 \\ \hline
\arrayrulecolor{black}\hline \hline
\arrayrulecolor{mygrey}
Nav. Stack & E7 & 25.2 min& 276.26 m & 0.18 m/s & 66\% & \bf 0/50 & \bf 12 & \bf 0 & \bf 17  \\ \hline
Goal Cont. (GC) & E7 & 45.9 min & 540.01 m & 0.19 m/s & 62\% & \bf 0/50 & 21 &  2 & 28  \\ \hline
Traj. Cont. (TC)  & E7 & 46.2 min & 416.81 m & 0.15 m/s & 64\% & 1/50 & 18 &  1 & 28 \\ \hline
Proposed System  & E7 & 39.4 min & 503.95 m & 0.21 m/s & \bf 84\% & \bf  0/50 & 16 & \bf 0 & 35 \\ 
\end{tabular}
}
\vspace{-10pt}
\end{table*}

\subsection{Evaluation Based on Objective Metrics}
\label{subsec:exp1}

\begin{description}[align=left,leftmargin=0em,labelsep=0.2em,font=\textbf,itemsep=0.5em]
\item[Experiments in Previously-Seen Maps.] 
We first conducted an evaluation on the Foyer map, which was used for training or tuning of algorithms, to evaluate the capacity of the models on a previously-seen environmental condition (E4) and in new conditions (E5 and E6). As indicated in Table \ref{tab:env}, we considered 100 test scenarios according to each of E4 and E5, and 50 test scenarios according to E6.

Table \ref{tab:results} presents quantitative results for this experiment. Overall, all methods  increased speed as the environments were simpler, suggesting that they adjusted navigation behavior based on environmental conditions. Unfortunately, though, collisions were observed throughout the experiment.

In terms of relative performance on the familiar condition E4, the proposed approach reached the goal the most. 
%
For new environmental conditions, the Navigation Stack was the best model in the simplest condition (E5), while the proposed approach was the best model in the more complex condition (E6). In E5, the performance of our system is worse than in the other two environments. We observed that in simulation, our model paid too much attention to the single pedestrian, and halted multiple times to let the pedestrian pass and as a result, could not reach the goal within 90 seconds. Remarkably, the performance of our approach varied only by $\pm{3\%}$ when reaching the goal in the foyer, while the performance of the Nav. Stack varied by about $\pm{8\%}$. It is difficult to manually find a fixed set of model parameters for all environmental conditions.

The GC and TC baselines showed their best performance in the simple condition E5, even in comparison to E4. This result suggests that the controllers were able to follow the global plan, but had trouble adapting to dynamic environments, which were more common in E4 and E6. 

\item[Generalization to a New Map.] 
We tested generalization capabilities in a new map of a university laboratory (environmental condition E7). The results are presented in the last four rows of Table \ref{tab:results}. Our proposed model reached the goal the most often, followed by the navigation stack. There is a marked difference in the goal reach percentages of the proposed system and the navigation stack. This result again highlights the difficulty in manually tuning static model parameters for navigation with ROS' navigation stack.

\end{description}



\subsection{Evaluation Based on Subjective Metrics}
\label{subsec:exphuman}
We surveyed 35 people through Amazon Mechanical Turk in regards to their perception of the motion of the robot in 48 videos of scenarios from environmental conditions E4, E5, E6, and E7 (Table \ref{tab:env}). We then ran REstricted Maximum Likelihood (REML) \cite{corbeil1976restricted} analyses on how Aggressive, Natural, and Efficient the motion looked with \textit{Method} (Nav. Stack, GC, TC, Proposed) as main effect and participant as random effect. The results for Efficiency were significant, F$[3, 1642]=19.26$, p$\,<0.01$. A Tukey post-hoc test \cite{abdi2010tukey} showed that the proposed navigation approach was perceived as significantly more efficient (M=$3.69$, SE=$0.06$) than the Nav. Stack and TC (M=$3.46$, SE=$0.07$; M=$3.05$, SE=$0.07$). The results for Aggressiveness and Naturalness were significant as well with p$\,<0.01$. The post-hoc showed that the TC baseline led to the most aggressive (M=$2.7$, SE=$0.06$) and least natural (M=$3.14$, SE=$0.06$) navigation behavior in comparison to all other methods. In particular, the ratings for our approach in terms of aggressiveness and naturalness were M=$2.30$, SE=$0.06$, and M=$3.62$, SE=$0.06$, respectively.

\subsection{Attention Visualization}
\begin{figure}[b]
    \centering
    \includegraphics[width=1\linewidth]{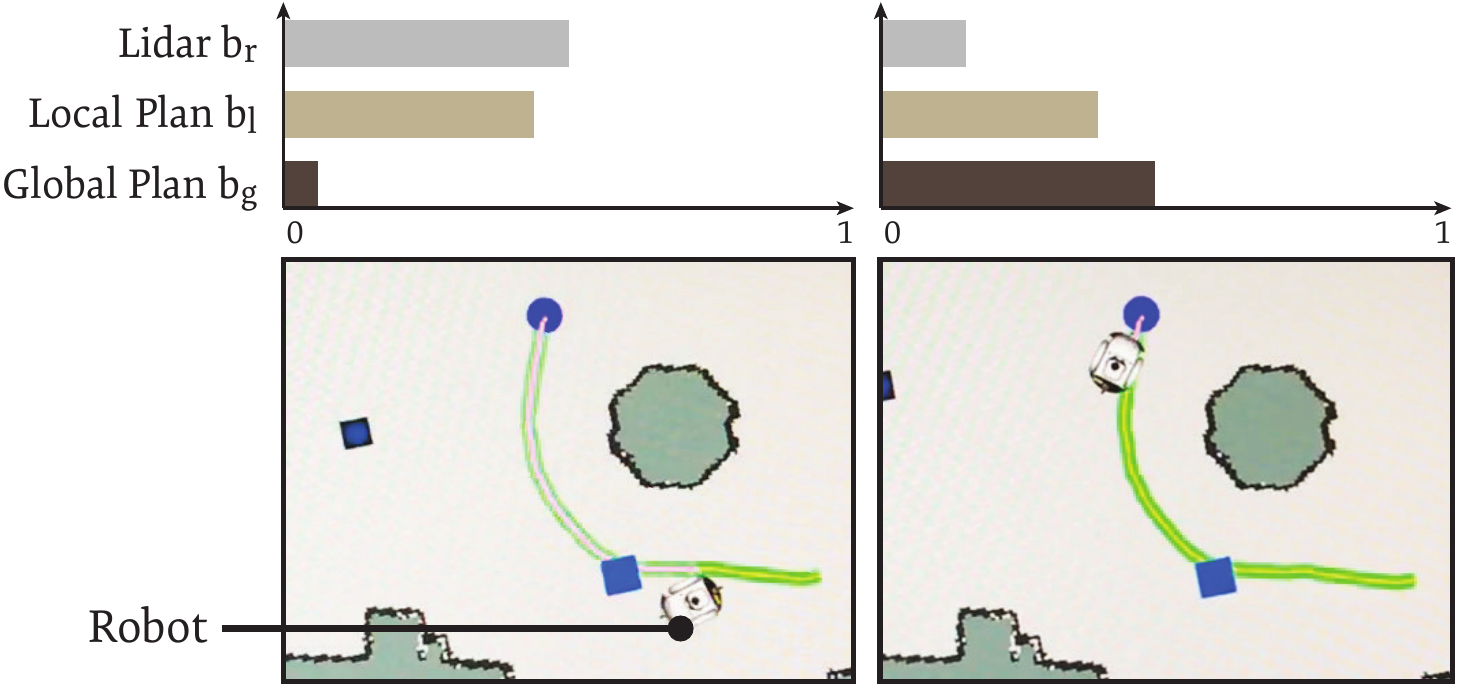}
    \caption{\small{Attention visualization for the Velocity Controller. Left: The robot navigates near an obstacle (blue square). Right: the robot approaches the goal (blue circle).}}
    \label{fig:attention}
    \vspace{-1em}
\end{figure}

We inspected the attention coefficients from the learned components of the proposed navigation system and found that they correlate consistently with specific task conditions, providing a mechanism to interpret the execution of the model.  For instance, as the robot navigated nearby obstacles, the attention coefficient $b_r$ for the lidar features of the Velocty Controller tended to grow, while the attention coefficient $b_g$ for the global plan features tended to get lower (left image of Fig. \ref{fig:attention}). Meanwhile, the opposite result was observed when the robot approached the goal (right image of Fig. \ref{fig:attention}). These results suggest that the proposed attention mechanisms helped the robot extract the right information from the many inputs to the components of our deep model. More visualizations can be seen in the supplementary video.

\section{Limitations}
We conducted experiments in relatively open simulated environments, but future work should also evaluate the performance of the proposed approach in more narrow spaces, like corridors. Further, it is important to conduct a systematic evaluation of our model in real-world scenarios. While we have attempted such preliminary tests and our system has worked effectively several times, it has also failed due to errors in people tracking and noise in real lidar measurements, e.g., due to material reflections, that were not typical of our simulation. We continue to work to address these issues.


\section{Conclusion}
\label{sec:dicussion}
We proposed a new navigation system that combines traditional planning with modern deep learning techniques. The learning components of the system were structured as a local planner and a velocity controller. This structure made our model interpretable. First, by visualizing the local plan, one could get a sense of how the robot would move and react to dynamic elements of the environment. Second, by inspecting the attention coefficients of our model, one could see the type of information that the robot was considering during navigation. Overall, our evaluation of the proposed model suggested that it was effective for navigation in complex dynamic environments. The performance of the proposed model was more consistent than the performance of a traditional two-layer navigation system with social costs~\cite{lu2014layered}. Moreover, the learning component of our approach was more successful at reaching the goal than other deep controllers that mapped sensor data and a global plan directly to velocity commands, e.g.,  \cite{pfeiffer2017perception}. Finally, our results reinforce the idea that imitation learning can facilitate modeling socially appropriate behavior from example navigation data \cite{shiarlis2017acquiring,tai2018social}.

\section*{Acknowledgments}
\label{sec:ack}
The Toyota Research Institute provided funds to assist with this research, but this paper solely reflects the opinions and conclusions of its authors and not of any Toyota entity. 



\bibliographystyle{IEEEtranN}
\footnotesize
\bibliography{references}

\end{document}